\documentclass[11pt,a4paper]{article}
\usepackage[hyperref]{acl2021}
\usepackage{times}
\usepackage{latexsym}

\usepackage{microtype}
\usepackage{algorithm2e}
\usepackage{amsfonts}
\usepackage{amsmath}
\usepackage{amssymb}
\usepackage{array}
\usepackage{booktabs}
\RequirePackage{pgfkeys,tikz}
\usepackage{ctable}
\usepackage{epstopdf}
\usepackage{etex}
\PassOptionsToPackage{pdftex}{graphicx}
\usepackage{listings} 
\usepackage{comment}
\usepackage{microtype}
\usepackage{multicol}
\usepackage{multirow}
\usepackage{paralist} 
\usepackage{pifont}
\usepackage{ragged2e}
\usepackage{setspace}
\usepackage{titletoc}
\usepackage[normalem]{ulem}
\PassOptionsToPackage{hyphens}{url}
\usepackage{wrapfig}
\usepackage{xcolor}
\usepackage{xspace}
\usepackage{subcaption}
\usepackage{float}

\RequirePackage{flushend}
\usetikzlibrary{shapes,positioning}
\RequirePackage[skip=.5ex plus 1ex]{caption}
\usepackage[backgroundcolor=yellow!30,linecolor=red, bordercolor=red,textsize=scriptsize]{todonotes}
\RequirePackage{adjustbox,array,multirow,colortbl}
\usepackage{color, colortbl}
\definecolor{Gray}{gray}{0.5}

\colorlet{newchanges}{black}
\newcommand\sys{{\sc CDNet}}

\aclfinalcopy 
\def\aclpaperid{4149} 

\title{Constraint based Knowledge Base Distillation\\ in End-to-End Task Oriented Dialogs}

\author{
Dinesh Raghu$^{\ *\ \dagger\ 1\ 2}$,
Atishya Jain$^{\ *\ 1}$,
Mausam$^{\ 1}$ and
Sachindra Joshi$^{\ 2}$\\
$^1$ IIT Delhi, New Delhi, India\\
$^2$ IBM Research, New Delhi, India\\
{\em \{diraghu1, jsachind\}@in.ibm.com,
\{Atishya.Jain.cs516, mausam\}@cse.iitd.ac.in}
}

\date{}

\begin{document}
\maketitle
\begin{abstract}
End-to-End task-oriented dialogue systems generate responses based on dialog history and an accompanying knowledge base (KB). Inferring those KB entities that are most relevant for an utterance is crucial for response generation. 
\textcolor{newchanges}{Existing state of the art scales to large KBs by softly filtering over irrelevant KB information. In this paper, we propose a novel filtering technique that consists of (1) a pairwise similarity based filter} that identifies relevant information by respecting the n-ary structure in a KB record. and, (2) an auxiliary loss that helps in separating contextually unrelated KB information.  \textcolor{newchanges}{We also propose a new metric -- multiset entity F1 which fixes a correctness issue in the existing entity F1 metric.} Experimental results on three publicly available task-oriented dialog datasets  show that our proposed approach outperforms existing state-of-the-art models.
\end{abstract}

\section{Introduction}

\begingroup
\renewcommand{\thefootnote}{\fnsymbol{footnote}}
\stepcounter{footnote}\footnotetext{D. Raghu and A. Jain contributed equally to this work.}
\stepcounter{footnote}\footnotetext{D. Raghu is an employee at IBM Research. This work was carried out as part of PhD research at IIT Delhi.}

\endgroup


Task oriented dialog systems interact with users to achieve specific goals such as restaurant reservation or calendar enquiry. To satisfy a user goal, the system is expected to retrieve necessary information from a knowledge base and convey it using natural language. 
Recently several end-to-end approaches \cite{BordesW16,wu2019globaltolocal,9053667,mem2seq} have been proposed \textcolor{newchanges}{for learning these dialog systems}.

\begin{figure}
\centering
\footnotesize
\includegraphics[scale=0.43]{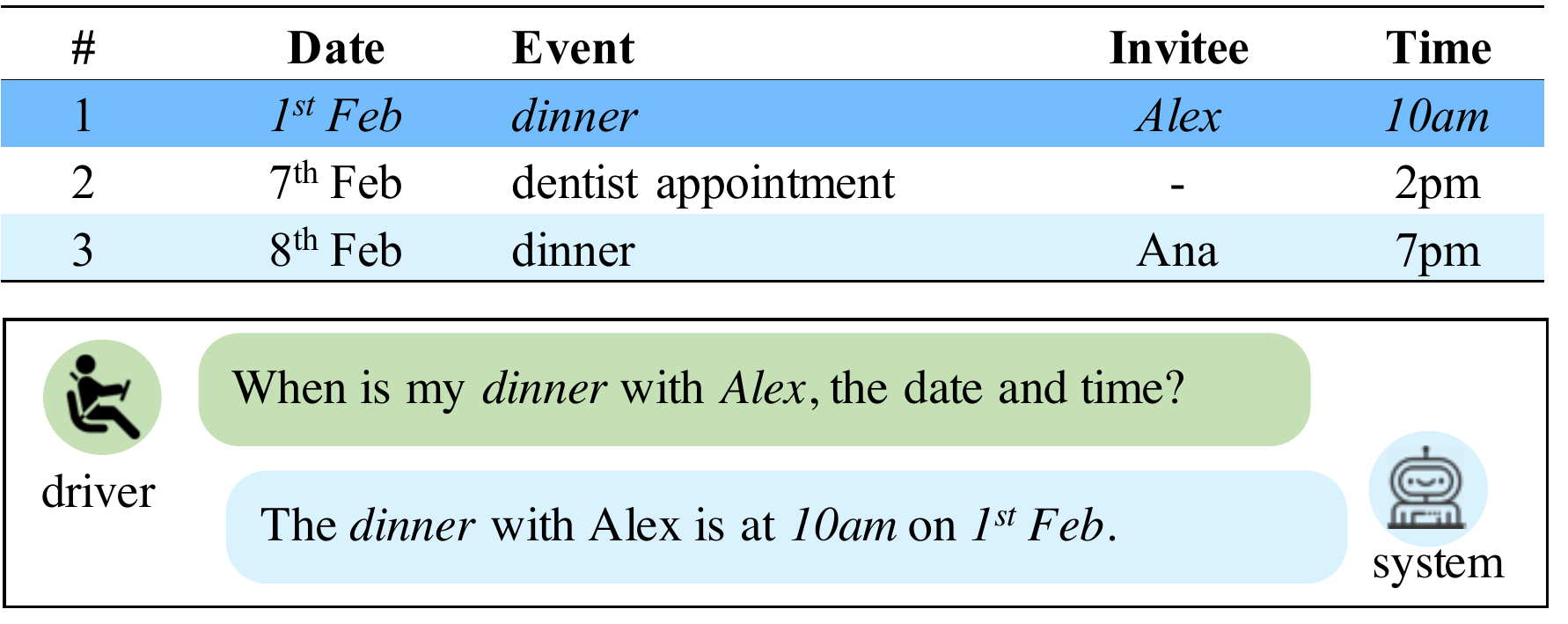}
\caption{An example dialog between a driver and a system along with the associated knowledge base.}
\label{fig:example}
\vspace{-.5cm}
\end{figure}

Inferring the most relevant KB entities necessary for generating the response is crucial for achieving task success. 
\textcolor{newchanges}{To effectively scale to large KBs, existing approaches \cite{wen-etal-2018-sequence,wu2019globaltolocal} distill the KB by softly filtering irrelevant KB information based on the dialog history. For example, in Figure \ref{fig:example} the ideal filtering technique is expected to filter just the row 1 as the driver is requesting information about \textit{dinner} with \textit{Alex}. But existing techniques often filter some irrelevant KB information along with the relevant KB information. For example, in Figure \ref{fig:example} row 3 may also get filtered along with row 1.}

Our analysis of \textcolor{newchanges}{the best performing distillation technique \cite{wu2019globaltolocal}} revealed that embeddings learnt for entities of the same type are quite close to each other. This may be due to entities of the same type often appearing in similar context in history and KB. Such embeddings hurt the overall performance as they reduce the gap between relevant and irrelevant KB records. For example, in Figure \ref{fig:example} row 3 may not get distilled out if \textit{Alex} and \textit{Ana} have similar embeddings. 

\textcolor{newchanges}{In this paper}, we propose Constraint based knowledge base Distillation NETwork (\sys), which (1) uses a novel pairwise similarity based distillation computation which distills KB at a record-level, and (2) an auxiliary loss which helps to distill contextually unrelated KB records by enforcing constraints on embeddings of entities of the same type. \textcolor{newchanges}{We noticed the popular entity F1 evaluation metric has a correctness issue when the response contains multi instances of the same entity value. To fix this issue, we propose a new metric called multiset entity F1.} We empirically show that \sys\ performs either significantly better than or comparable to existing approaches on three publicly available task oriented dialog datasets.

\begin{figure*}[ht]
\centering
\footnotesize
\includegraphics[scale=0.45]{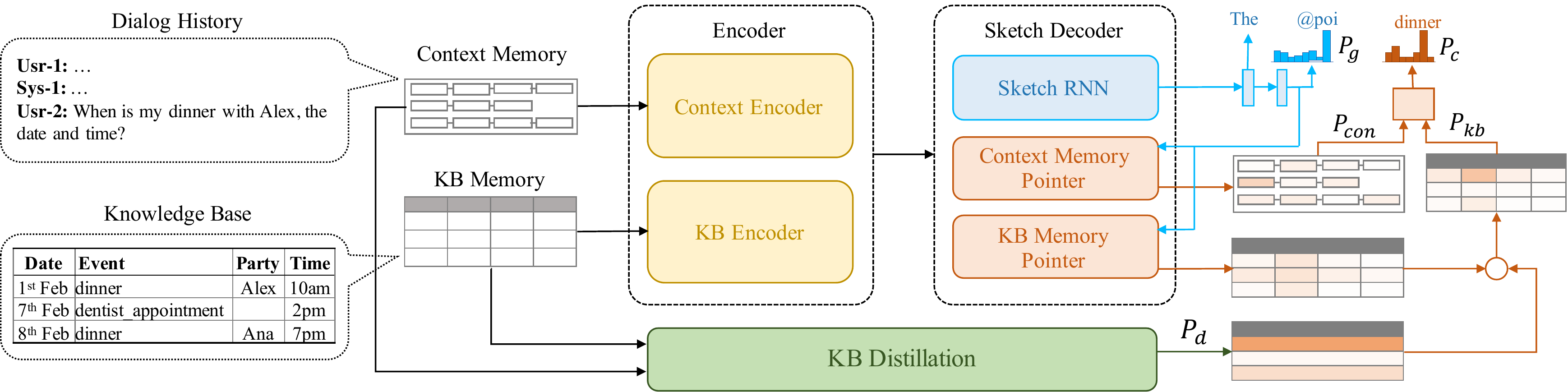}
\caption{Architecture of \sys\ model.}
\label{fig:system}
\end{figure*}

\section{Related Work}
We first discuss approaches that are closely related to our work.
\textcolor{newchanges}{\newcite{wu2019globaltolocal} perform KB distillation but fails to capture the relationship across attributes in KB records. It represents a KB record with multiple attributes as a set of triples (\textit{subject, predicate, object}). This breaks direct connection between record attributes and requires the system to reason over longer inference chains. In Figure \ref{fig:example}, if \textit{event} field is used as the key to break the record into triples, then the distillation has to infer that \textit{(dinner, invitee, Alex)}, \textit{(dinner, date, 1$^{st}$ Feb)} and \textit{(dinner, time, 10am)} are connected. In contrast, \sys\ performs KB distillation by maintaining the attribute relationships.
\newcite{wen-etal-2018-sequence} perform distillation using the similarity between dialog history representation and each attribute representation in a KB record, whereas \sys\ uses word based pairwise similarity for distillation.}

We now briefly discuss approaches that improve other aspects of task oriented dialogs. \newcite{He2020} and \newcite{9053667} model KBs using Relational GCNs \cite{schlichtkrull2017modeling}.\textcolor{newchanges}{\newcite{raghu-etal-2019-disentangling} provide support for entities unseen during train. \newcite{reddy2018multilevel} improve the ability to reason over KB by respecting the relationships between connected attributes.} \newcite{qin2019entityconsistent} restricts the response to contain entities from a single KB record. \cite{qin-etal-2020-dynamic} handle multiple domains using shared-private networks and \newcite{he-etal-2020-amalgamating} optimize their network on both F1 and BLEU. We are the first to propose a pairwise similarity score for KB distillation and a embedding constraint loss to distill irrelevant KB records. 


\section{\sys}
\sys\ \footnote{\url{https://github.com/dair-iitd/CDNet}} has an encoder-decoder architecture that takes as input (1) the dialog history $\mathcal{H}$, modelled as a sequence of utterances $\{u_i\}_{i=1}^{k}$ and each utterance $u_i$ as sequence of words $\{w_{i}^j\}$, and (2) a knowledge base $\mathcal{K}$ with $M$ records $\{r_m\}_{m=1}^M$ and each record $r_m$ has $N$ key-value attribute pairs $\{(k^n, v_{m}^n)\}_{n=1}^N$.
The network generates the system response 
$Y=(y_1, \ldots, y_{\mathbb{T}})$ one word at a time.

\subsection{\sys\ Encoder}
\noindent \textbf{Context Encoder:} The dialog history $\mathcal{H}$ is encoded using a hierarchical encoder \cite{sordoni2015hierarchical}. Each utterance representation $\mathbf{u}_i$ is computed using a Bi-GRU \cite{schuster1997bidirectional}. We denote the hidden state of the $j^{th}$ word in $i^{th}$ utterance as $\mathbf{w}_{i}^j$. The context representation $\mathbf{c}$ is generated by passing $\mathbf{u}_i$s through a GRU.

\vspace{0.5ex}
\noindent \textbf{KB Encoder:} We encode the KB using the multi-level memory proposed by \newcite{reddy2018multilevel} as its structure allows us to perform distillations over KB records. The KB memory contains two-levels. The first level is a set of KB records. Each KB record is represented as sum of its attributes $\mathbf{r}_m = \sum\nolimits_{n}\Phi^{e}(v_{m}^n)$, where $\Phi^{e}$ is the embedding matrix. In the second level, each record is represented as a set of attributes. Each attribute is a key-value pair, where the key $\mathbf{k}^n$ is the attribute type embedding and the value $\mathbf{v}_{m}^n$ is the attribute embedding.

\subsection{KB Distillation}

The KB distillation module softly filters irrelevant KB records based on the dialog history by computing a distillation distribution ($P_d$) over the KB records. To compute $P_d = [d_1, \ldots, d_M]$, we first score each KB record $r_m$ based on the dialog history $\mathcal{H}$ as follows:

\begin{gather}
s_m = \sum_{w \in \mathcal{H}} \sum_{v^n_m \in r_m} \text{CosSim}(\Phi^{e}(w), \Phi^{e}(v^n_m))
\end{gather}

\noindent where CosSim is the cosine similarity between two vectors. The distillation likelihood $d_m$ for each record ${r}_m$ then is given by $ d_m = \text{exp}(s_m)/\sum\nolimits_{q=1}^{M} \text{exp}(s_q)$.

Defining distillation distribution over the KB records rather than KB triples has two main advantages: (1) attributes (such as invitee, event, time and date in Figure \ref{fig:example}) in a KB record are directly connected and thus easy to distill, (2) it helps to distill the right records even when the record keys are not unique. In Figure \ref{fig:example}, row 3 would be distilled even though it shares the same event name.

\subsection{\sys\ Decoder}

Following \newcite{wu2019globaltolocal}, we first generate a sketch response which uses entity type (or sketch) tag in place of an entity. For example, \textit{The @meeting with @invitee is at @time} is generated instead of \textit{The dinner with Alex is at 10pm}. When an entity tag is generated, we choose an entity suggested by the context and KB memory pointers. 

\vspace{0.5ex}
\noindent \textbf{Sketch RNN:} We use a GRU to generate the sketch response. At each time $t$, a generate distribution $P_{g}$ is computed using the decoder hidden state $\mathbf{h}_t$ and an attended summary of the dialog context $\mathbf{g}_t$. The summary $\mathbf{g}_t = \sum\nolimits_i \sum\nolimits_j a_{ij} \mathbf{w}_{i}^j$, where $a_{ij}$ is the Luong attention \cite{luong2015effective} weights over the context word representations ($\mathbf{w}_{i}^j$).

\vspace{0.5ex}
\noindent \textbf{Context Memory Pointer:} At each time $t$, generate the copy distribution over the context $P_{con}$ by performing a multi-hop Luong attention over the context memory. The initial query $\mathbf{q}_t^0$ is set to $\mathbf{h}_t$. $\mathbf{q}_t^0$ is then attended over the context to generate an attention distribution $\mathbf{a}^1$ and a summarized context $\mathbf{g}_t^1$. We represent this as $\mathbf{g}_t^1 = \text{Hop}(\mathbf{q}_t^0, \mathbf{x})$. In the next hop the same process is repeated by updating the query $\mathbf{q}_t^1=\mathbf{q}_t^0+\mathbf{g}_t^1$. The attention weights after $H$ hops is used for computing the context pointer $P_{con}$ as follows:
\begin{gather}
P_{con}(y_t=w) = \sum\nolimits_{ij:w_{i}^j=w} a_{ij}^H 
\label{eq:pcon} 
\end{gather}

\noindent \textbf{KB Memory Pointer:} At each time $t$, we generate the copy distribution over the KB $P_{kb}$ using (1) Luong attention weight $\beta_m^t$ over the KB record $r_m$ and (2) Luong attention weight $\gamma_n^t$ over attribute keys in a record $\mathbf{k}^n$ and (3) the distillation weight $d_m$ over the KB record $r_m$. The KB pointer $P_{kb}$ is computed as follows:
\begin{align}
P_{kb}(y_t=w) = \frac{\sum\nolimits_{mn:v_m^n=w} d_m \beta_m^t \gamma_n^t}{\sum\nolimits_{mn} d_m \beta_m^t \gamma_n^t} 
\end{align}
The two copy pointers are combined using a soft gate $\alpha$ \cite{see2017point} to get the final copy distribution $P_c$ as follows, 

\begin{gather}
    P_{c}(y_t) = \alpha P_{kb}(y_t) + (1-\alpha) P_{con}(y_t)
\end{gather}

\subsection{Loss}
We guide the distillation module using two auxiliary loss terms: entity constraint loss $\mathcal{L}_{ec}$ and distillation loss $\mathcal{L}_{d}$. Often entities of the same type (e.g., \textit{Ana} and \textit{Alex}) have embeddings similar to each other. As a result, records with similar but unrelated entities are incorrectly assigned a high distillation likelihood. To alleviate this problem, we make the cosine similarity between two entities of the same type to be as low as possible. This is captured by the constraint loss $\mathcal{L}_{ec}$ given by,
\begin{equation}
\mathcal{L}_{ec} = \sum\noindent_{(e_a, e_b) \in \mathcal{E}} \text{CosSim}(\Phi^{e}(e_a),\Phi^{e}(e_b))
\end{equation}
\noindent where $\mathcal{E}$ is a set of entity pairs in the KB that belong to the same entity type. 

The distillation likelihood $d_m$ of a KB record $r_m$ depends on the similarity between entities in the record and the words mentioned in the dialog context. We compute the distillation loss by defining reference distillation distribution $d_m^{*}$ as $s_m^*/\sum_{q=1}^{M} s_q^*$, where $s_m^*$ is the number of times any attribute in $r_m$ occurs in $\mathcal{H}$ and in the gold response. The distillation loss is given by, 
\begin{align}
\mathcal{L}_{d} = - \sum\nolimits_{m=1}^{M} d_m^{*}  \text{log}(d_m)
\end{align}

\noindent The overall loss function $\mathcal{L} = \mathcal{L}_{g} + \mathcal{L}_{c} + \mathcal{L}_{ec} + \mathcal{L}_{d}$, where $\mathcal{L}_{g}$ and $\mathcal{L}_{c}$ are the cross entropy loss on $P_g$ and $P_c$ respectively. Detailed equations are described in Appendix \ref{app-sec:detailed-equations}. 

\section{Experimental Setup}

\noindent{\textbf{Datasets:}}
We evaluate our model on three datasets -- CamRest \cite{wen2017network}, Multi-WOZ 2.1 (WOZ) \cite{budzianowski2018multiwoz} and Stanford Multi-Domain (SMD) Dataset \cite{Ericsigdial}. 

\vspace{0.5ex}
\begin{table*}[h]
\centering
\footnotesize
\begin{tabular}{c|l|c|c}
\toprule
& \textbf{Utterance} & \textbf{Set} & \textbf{MultiSet} \\
 \midrule
 \midrule

\multirow{2}{*}{\textbf{Gold}} & for which one? I have two, one on the \textit{8th} at \textit{11am} and & \textit{\{8th, 11am, wednesday\}} & \textit{\{8th, 11am, wednesday, 11am\}} \\ & one on \textit{wednesday} at \textit{11am} & & \\
\midrule
\textbf{Pred-1} & your appointment is on \textit{8th} at \textit{11am} & \textit{\{8th, 11am\}} & \textit{\{8th, 11am\}} \\
\midrule
\textbf{Pred-2} & your appointment is on \textit{8th} on \textit{8th} on \textit{8th} and on \textit{8th} & \textit{\{8th\}} & \textit{\{8th 8th, 8th, 8th\}} \\
\bottomrule
\end{tabular}
\caption{An example to demonstrate the correctness issue with the Entity F1 metric.}
\label{tab:newmetric}
\end{table*}

\begin{table*}[h]
\centering
\footnotesize
\begin{tabular}{l|ccc|ccc|ccc}
\toprule
& \multicolumn{3}{c|}{\textbf{CamRest}} & \multicolumn{3}{c|}{\textbf{SMD}} & \multicolumn{3}{c}{\textbf{WOZ 2.1}}  \\
\cmidrule{2-10}
\textbf{Model} & \textbf{BLEU} & \textbf{\textcolor{Gray}{Ent. F1}} & \textbf{MSE F1} &  \textbf{BLEU} & \textbf{\textcolor{Gray}{Ent. F1}} & \textbf{MSE F1} & \textbf{BLEU} & \textbf{\textcolor{Gray}{Ent. F1}} & \textbf{MSE F1} \\
\midrule
DSR \cite{wen-etal-2018-sequence} & 18.3 & \textcolor{Gray}{53.6} & - & 12.7 & \textcolor{Gray}{51.9} & - & 9.1 & \textcolor{Gray}{30.0} & - \\
GLMP \cite{wu2019globaltolocal} & 15.1 & \textcolor{Gray}{58.9} & 57.5 & 13.9 & \textcolor{Gray}{59.6} & 59.6 & 6.9 & \textcolor{Gray}{32.4} & - \\
MLM \cite{reddy2018multilevel} & 15.5 & \textcolor{Gray}{62.1} & - & 17 & \textcolor{Gray}{54.6} & - & - & \textcolor{Gray}{-} & - \\
Ent. Const. \cite{qin2019entityconsistent} & 18.5 & \textcolor{Gray}{58.6} & - & 13.9 & \textcolor{Gray}{53.7} & - & - & \textcolor{Gray}{-} & - \\
TTOS \cite{he-etal-2020-amalgamating} & 20.5 & \textcolor{Gray}{61.5} & - & 17.4 & \textcolor{Gray}{55.4} & - & - & \textcolor{Gray}{-} & - \\
DFNet \cite{qin-etal-2020-dynamic} & - & \textcolor{Gray}{-} & - & 14.4 & \textcolor{Gray}{62.7} & 56.7 & 9.4 & \textcolor{Gray}{35.1} & 34.8 \\
EER \cite{He2020} & 19.2 & \textcolor{Gray}{65.7} & 65.5 & 17.2 & \textcolor{Gray}{59.0} & 55.1 & 13.6 & \textcolor{Gray}{35.6} & 35.0 \\
FG2Seq \cite{9053667} & 20.2 & \textcolor{Gray}{66.4} & 65.4 & 16.8 & \textcolor{Gray}{61.1} & 59.1 & \textbf{14.6} & \textcolor{Gray}{36.5} & 36.0 \\
\midrule
\sys\ & \textbf{21.8} & \textbf{\textcolor{Gray}{68.6}} & \textbf{68.4} & \textbf{17.8} & \textbf{\textcolor{Gray}{62.9}} & \textbf{62.9} & 11.9 & \textbf{\textcolor{Gray}{38.7}} & \textbf{38.6} \\
\bottomrule
\end{tabular}
\caption{Performance of \sys\ and baselines on the CamRest, SMD and Multi-WOZ 2.1 datasets.}
\label{tab:smd}
\end{table*}

\noindent{\textbf{Baselines:}} 
We compare \sys\ against the following baselines: MLM \cite{reddy2018multilevel}, DSR \cite{wen-etal-2018-sequence}, GLMP \cite{wu2019globaltolocal}, Entity Consistent \cite{qin2019entityconsistent}, EER \cite{He2020}, FG2Seq \cite{9053667}, TTOS \cite{he-etal-2020-amalgamating} and DFNet \cite{qin-etal-2020-dynamic}.

\vspace{0.5ex}

\noindent \textbf{Training Details:}
\sys\ is trained end to end using Adam optimizer \cite{kingma2014adam}.
The embedding dimensions of the hidden states of encoder and decoder GRU are set to 200 and 100 respectively. Word embeddings are initialized with pre-trained 200d GloVe embeddings \cite{glovevec}. Words not in Glove are initialized using Glorot uniform distribution \cite{pmlr-v9-glorot10a}. The dropout rate is set to 0.2 and teacher forcing ratio set to 0.9. The best hyper-parameter setting for each dataset and other training details are reported in the Appendix \ref{appendix:hyperparamters}.

\vspace{0.5ex}
\noindent \textbf{Evaluation Metrics:} \textcolor{newchanges}{We measure the performance of all the models using BLEU \cite{papineni2002bleu}, our proposed multiset entity F1 and for completeness the previously used entity F1 \cite{wu2019globaltolocal}.}

\vspace{0.5ex}
\noindent \textbf{MultiSet Entity F1 (MSE F1):} 
\textcolor{newchanges}{The entity F1 is used to measure the model's ability to predict relevant entities from the KB. It is computed by micro averaging over the set of entities in the gold responses and the set of entities in the predicted responses. This metric suffers from two main problems. First, when the gold response has multiple instances of the same entity value, it is accounted for just once in the set representation. For example, in Table \ref{tab:newmetric} the entity value \textit{11am} occurs twice in the gold response but accounted for just once in the {\em set} representation. As a result the recall computation does not penalize the prediction \textit{pred-1} for missing an instance of \textit{11am}. Second, the existing metric fails to penalize models that stutter. For example,  in Table \ref{tab:newmetric} the precision of \textit{pred-2} is not penalized for repeating the entity value \textit{8th}. We propose a simple modification to the entity F1 metric to fix these correctness issues. The modified metric, named MultiSet Entity F1, is computed by micro averaging over the {\em multiset} of entities rather than a {\em set}. As multisets allow multiple instances of same entity values, it  (1) accounts for the same entity value mentioned more than once in the gold by penalizing recall for missing any instances and (2) accounts for models that stutters by penalizing the precision.}



\section{Results}
\label{sec:results}


The results are shown in Table \ref{tab:smd}.  On CamRest and SMD, \sys\ outperforms existing models in both MSE F1 and BLEU. On WOZ, \sys\ achieves best only in MSE F1. We observed that the responses generated by \sys\ on WOZ were appropriate, but did not have good lexical overlap with the gold responses. To investigate this further, we perform a human evaluation of the responses predicted by \sys, FG2Seq and EER.

\begin{figure*}[ht]
\centering
\footnotesize
\subcaptionbox{\label{exp1:a}}{\includegraphics[scale=0.23]{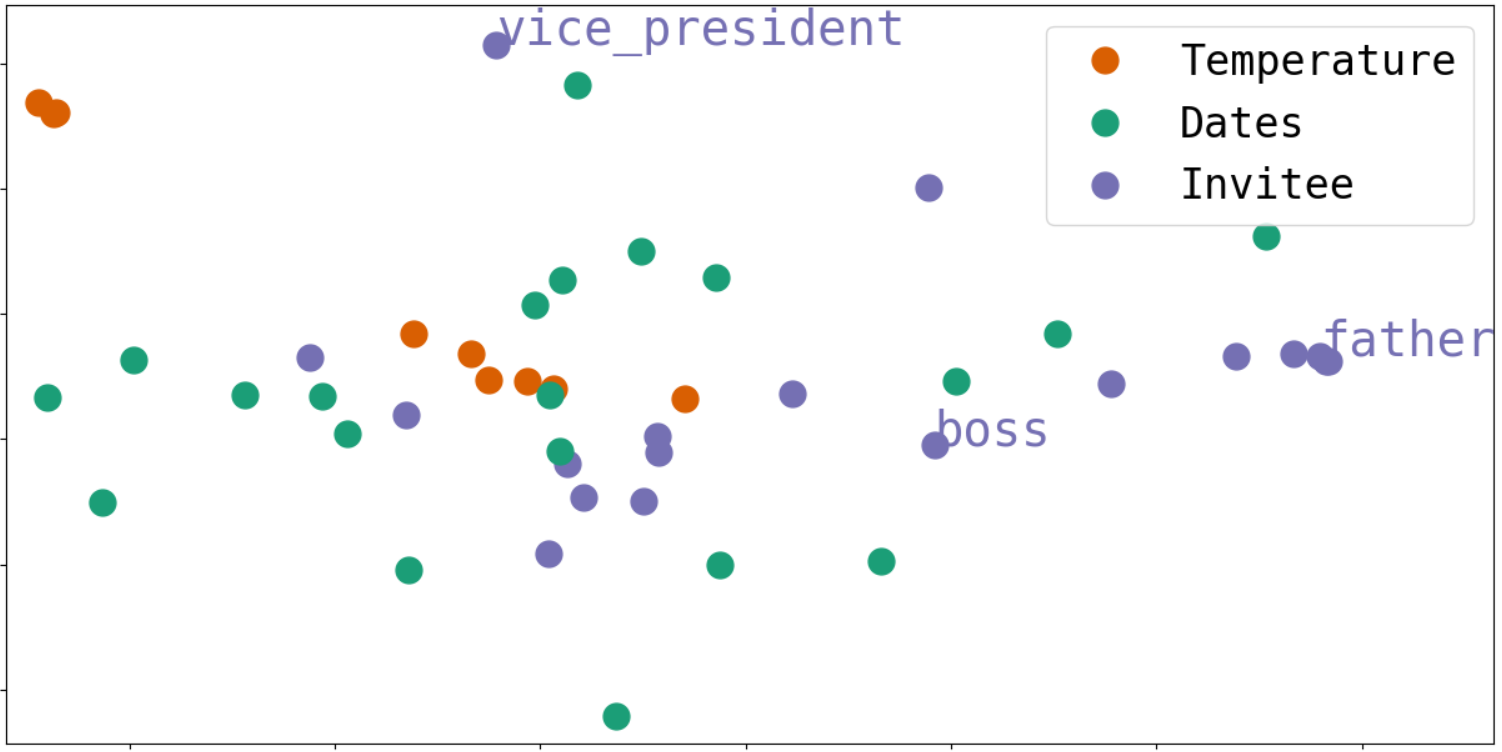}}
\subcaptionbox{\label{exp1:b}}{\includegraphics[scale=0.23]{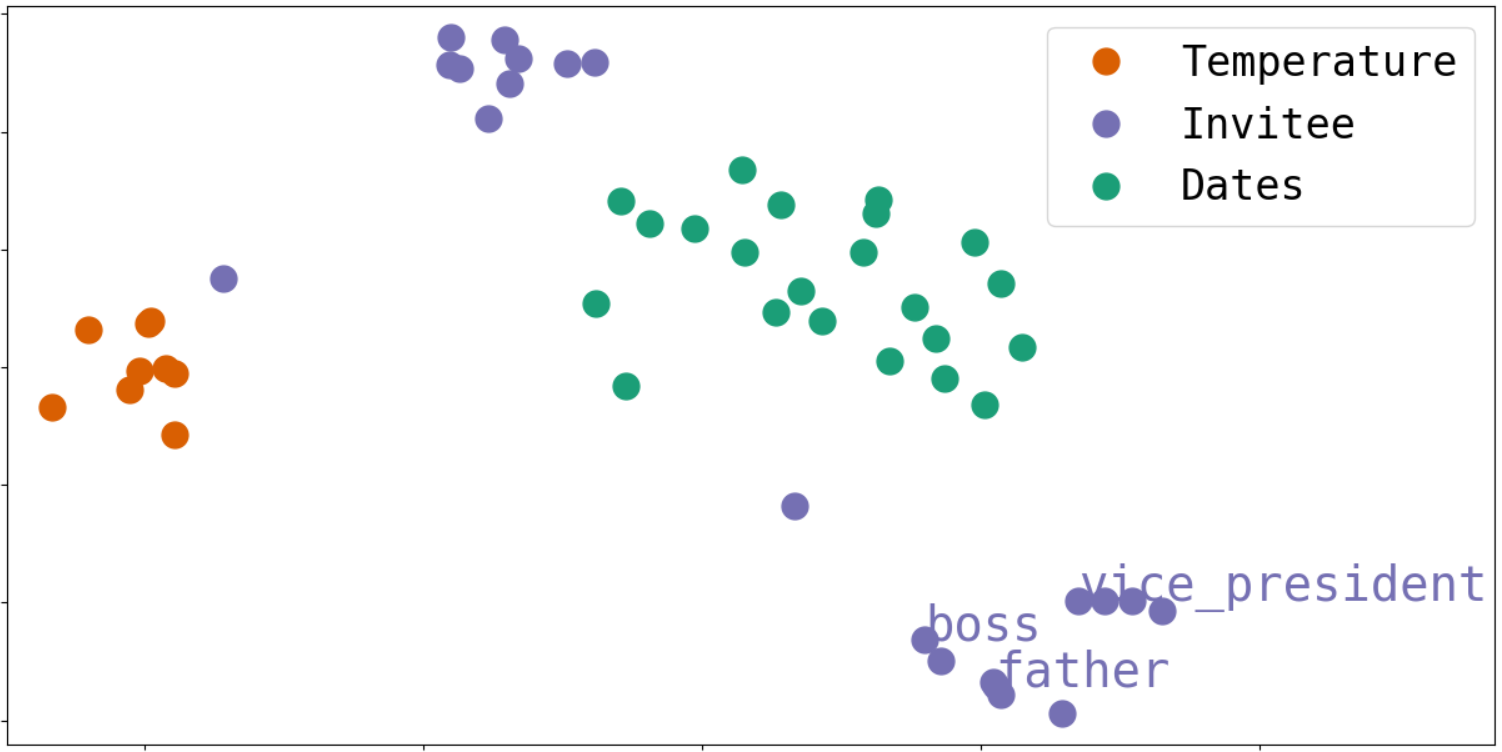}}

\caption{T-sne plots of entity embeddings from SMD of (a) \sys\ \& (b) GLMP.}
\label{fig:exp1}
\end{figure*}

\vspace{0.5ex}
\noindent \textbf{Human Evaluation:} We conduct a human evaluation to assess two dimensions of generated responses: (1) \textit{Appropriateness}: how useful are the responses for the given dialog context and KB, and (2) \textit{Naturalness:} how human-like are the predicted responses. We randomly sampled 75 dialogs from each of the three datasets and requested two judges to evaluate on a Likert scale \cite{likert1932technique}. The results are summarized in Table \ref{tab:hev}. \sys\ outperforms both FG2Seq and EER on \textit{appropriateness} across all three datasets. Despite having a lower BLEU score on WOZ, \sys\ performs in-par with the other two baselines on \textit{naturalness}.

\begin{table}[t]
\centering
\footnotesize
\begin{tabular}{c|ccc|ccc}
\toprule
& \multicolumn{3}{c|}{\textbf{Appropriateness}} & \multicolumn{3}{c}{\textbf{Naturalness}}  \\
\cmidrule{2-7}
\textbf{Model} & \textbf{SMD} & \textbf{Cam} & \textbf{WoZ} & \textbf{SMD} & \textbf{Cam} & \textbf{WoZ} \\
\midrule
EER & 2.9 & 3.8 & 3.4 & 3.6 & 4.2 & 4.0 \\
FG2Seq & 3.1 & 3.7 & 3.7 & 3.9 & 4.3 & 4.0 \\
\sys\ & 3.6 & 4.1 & 3.9 & 3.7 & 4.3 & 4.1 \\
\bottomrule
\end{tabular}
\caption{Human Evaluation of \sys\ on the CamRest, SMD and Multi-WOZ 2.1 datasets.}
\label{tab:hev}
\end{table}

\vspace{0.5ex}
\noindent \textbf{Ablation Study:} We perform an ablation study by defining three variants. Table \ref{tab:abl} shows the MSE F1 and BLEU for the two settings on CamRest and SMD datasets. (1) We remove the entity constraint loss $\mathcal{L}_{ec}$ from the overall loss $\mathcal{L}$. (2) We replace our pairwise similarity based score $s_m$ used for KB distillation with the global pointer score ($s_m=\mathbf{r}_m^T.\mathbf{c}$) proposed by \cite{wu2019globaltolocal}. We refer to this setting as naive distillation. (3) We replace our pairwise similarity based score $s_m$ with the entry-level attention proposed by \cite{wen-etal-2018-sequence}. We see that both our contributions: pairwise similarity scorer for computing distillation distribution and the entity constraint loss contribute to the overall performance.

\begin{table}[t]
\centering
\footnotesize
\begin{tabular}{c|cc|cc}
\toprule
& \multicolumn{2}{c|}{\textbf{CamRest}} & \multicolumn{2}{c}{\textbf{SMD}}  \\
\cmidrule{2-5}
\textbf{Model} & \textbf{BLEU} & \textbf{MSE F1} &  \textbf{BLEU} & \textbf{MSE F1} \\
\midrule
\sys\ & \textbf{21.8} & \textbf{68.4} & \textbf{17.8} & \textbf{62.9}\\
\midrule
No $\mathcal{L}_{ec}$ & 19.2 & 65.4 & 17.4 & 62.2 \\
Naive Dist. & 15.0 & 64.2 & 16.9 & 60.6 \\
Entry-Level Attn. & 16.2 & 62.0 & 17.1 & 59.4 \\
\bottomrule
\end{tabular}
\caption{Ablation study of \sys\ on the CamRest and SMD datasets.}
\label{tab:abl}
\vspace{-.5cm}
\end{table}



\vspace{0.5ex}
\noindent \textbf{Discussion:}
We now discuss the effect of the entity constraint loss $\mathcal{L}_{ec}$ on the KB entity embeddings. Figure \ref{fig:exp1} shows the t-SNE plot \cite{van2008visualizing} of entity embeddings of \sys\ and GLMP where entities of the same type are represented using the same colour. We see that entities of the same type (e.g. \textit{father} and \textit{boss} of the type \textit{invitees}) are clustered together in embedding space of GLMP, while they are distributed across the space in \sys. This shows that the entity constraint loss has helped reduce the embedding similarity between entities of the same type and ensures KB records with similar but unrelated entities are filtered by the KB distillation. Visualization of distillation distribution helping identify relevant KB entities is shown in Appendix \ref{app-sec:distillation-viz}.



\section{Conclusion}
We propose \sys\ for learning end-to-end task oriented dialog system. \sys\ performs KB distillation at the level of KB records, thereby respecting the relationships between the connected attributes. \sys\ uses a pairwise similarity based score function to better distill the relevant KB records. By defining constraints over embeddings of entities of the same type, \sys\ filters out contextually unrelated KB records. \textcolor{newchanges}{We propose a simple modification to the entity F1 metric that helps fix correctness issues. We refer to the new metric as multiset entity F1. \sys\ significantly outperforms existing approaches on multiset entity F1 and \textit{appropriateness}, while being comparable on \textit{naturalness} and BLEU. We release the code for further research.}

\section*{Acknowledgments}
This work is supported by IBM AI Horizons Network grant, an IBM SUR award, grants by Google, Bloomberg and 1MG, a Visvesvaraya faculty award by Govt. of India, and the Jai Gupta chair fellowship by IIT Delhi. We thank the IIT Delhi HPC facility for computational resources.



\bibliographystyle{acl_natbib}
\bibliography{anthology,acl2021}

\clearpage
\appendix

\section{Training Details:}
\label{appendix:hyperparamters}
All the hyper parameters are finalised after a grid search over the dev set. We sample learning rates (LR) from $\{2.5\times10^{-4}, 5\times10^{-4}, 10^{-4}\}$. The Disentangle Label Dropout (DLD) rate \cite{raghu-etal-2019-disentangling} is sampled from $\{0.0, 0.05, 0.10, 0.15, 0.20\}$. The number of hops H in the response decoder is sampled from $\{1, 3, 5\}$.  We ran each hyperparameter setting 10 times and use the setting with the best validation entity F1. The best performing hyperparameters for all datasets are listed in Table \ref{tab:hyp}. 

\begin{table}[ht]
\centering
\begin{tabular}{c|c|c|c|c}
\toprule
\multicolumn{1}{c|}{\textbf{Dataset}} & \multicolumn{1}{c|}{\textbf{Hops}} & \multicolumn{1}{c|}{\textbf{DLD}} & \multicolumn{1}{c|}{\textbf{ LR }} & \multicolumn{1}{c}{\textbf{Val MSE F1}} \\
\cmidrule{1-5}
CamRest & 1 & 0\% & 0.0005 & 68.6 \\
SMD & 3 & 5\% & 0.00025 & 60.4\\
WoZ 2.1 & 3 & 0\% & 0.00025 & 34.3\\
\bottomrule
\end{tabular}
\caption{Best performing hyperparameters along with the best validation Entity F1 (Val Ent. F1) achieved for the three datasets.}
\label{tab:hyp}
\end{table}

All experiments were run on a single Nvidia V100 GPU with 32GB of memory. \sys\ has an average runtime of 3 hours (6 min per epoch), 10 hours (20 min per epoch) and 24 hours (36 min per epoch) on CamRest, SMD and WOZ respectively. \sys\ has a total of 2.8M trainable parameters (400K for embedding matrix, 720K for context encoder, 240k for the sketch RNN and 1440k for the Memory pointers). 

\section{Detailed Equations}
\label{app-sec:detailed-equations}
In this section, we describe the details of context encoder, \sys decoder and the loss.
\subsection{Context Encoder}
Given a dialog history $\mathcal{H}$ we compute the utterance representation $\mathbf{u}_i$ and context representation $\mathbf{c}$ as follows:
\begin{gather}
\mathbf{u_i} = \text{BiGRU}(\Phi^e(w_{i}^1), \ldots, \Phi^e(w_{i}^{\tau_i})) \\
\mathbf{c} = \text{GRU}(\mathbf{u}_{1}, \ldots, \mathbf{u}_{k})
\end{gather}
where $\tau_i$ is the number of words in utterance $u_i$ and $w_{i}^j$ is the $j^{th}$ word in the $i^{th}$ utterance.

\subsection{\sys\ Decoder}

Let $\mathbf{h_{t}}$ and $y_{t}$ be the hidden state and the predicted word at time $t$ respectively. The hidden state is computed as follows,
\begin{equation}
\mathbf{h}_t = \text{GRU}(\Phi^{e}(y_{t-1}), \mathbf{h}_{t-1})
\end{equation}
Now, we compute multi-hop Luong attention over the words representations $\mathbf{w}_{i}^j$ in the context memory. We set the initial query $\mathbf{q_t^0}$ to $\mathbf{h_t}$ and then apply Luong attention as follows:
\begin{equation}
a_{ij}^1 = \frac{\text{exp}(W_1\text{tanh}(W_2[\mathbf{w}_{i}^j, \mathbf{q}_t^0]))}{\underset{ij}{\Sigma}\text{exp}(W_1\text{tanh}(W_2[\mathbf{q}_t^0, \mathbf{w}_{i}^j]))}
\end{equation}
where $W_1$, $W_2$ are trainable parameters.
We then compute the summarized context representation $\mathbf{g}_t^1$ and the next hop query as follows:
\begin{gather}
\mathbf{g}_{t}^1 = \underset{ij}{\Sigma}a_{ij}^1\mathbf{w}_{i}^j \\
\mathbf{q}_{t}^{1} = \mathbf{q}_t^0 + \mathbf{g}_t^1
\end{gather}
We repeat this for $H$ hops. The attention vector after $H$ hop is represented $\mathbf{a}^H$. The generate distribution $P_g$ is given by:
\begin{equation}
    P_{g} = \text{Softmax}(W_3[\mathbf{h}_t, \mathbf{g}_t^1] + b_1)
\end{equation}
where $W_3$ and $b_1$ are trainable parameters. The context copy distribution $P_{con}$ is computed as follows:
\begin{gather}
P_{con}(y_t=w) = \sum\nolimits_{ij:w_{i}^j=w} a_{ij}^H 
\end{gather}

\noindent The KB copy distribution $P_{kb}$ is given by,
\begin{gather}
\beta_m^t = \text{Softmax}(W_4\text{tanh}(W_5[\mathbf{g}_t^H, \mathbf{h}_t, \mathbf{r}_{m}])) \\
\gamma_n^t = \text{Softmax}(W_6\text{tanh}(W_7[\mathbf{g_t}^H, \mathbf{h}_t, \mathbf{k}^{n}])) \\
P_{kb}(y_t=w) = \frac{\sum\nolimits_{mn:v_m^n=w} d_m \beta_m^t \gamma_n^t}{\sum\nolimits_{mn} d_m \beta_m^t \gamma_n^t} 
\end{gather}
where $W_4$, $W_5$, $W_6$ and $W_7$ are trainable parameters.
Now we compute the gate $\alpha$ to combine $P_{con}$ and $P_{kb}$ to get a final copy distribution $P_c$ as follows:
\begin{gather}
    \mathcal{K}_t = \sum_{m}d_m \beta_m^t \mathbf{r}_m \\
    \alpha = \text{Sigmoid}(W_8[\mathbf{h}_t,\mathbf{g}_t^H,\mathcal{K}_t]) \\
    P_{c} = \alpha P_{kb}(y_t) + (1-\alpha) P_{con}(y_t)
\end{gather}
where $W_8$ is a trainable parameter.

\subsection{Loss}
We compute the cross entropy loss over the generate $P_g$ and copy $P_c$ distribution as follows:
\begin{gather}
\mathcal{L}_g = - \sum_{t=1}^{T}\text{log}(P_g(y_t)) \\
\mathcal{L}_c = - \sum_{t=1}^{T}\text{log}(P_c(y_t))
\end{gather}

\section{Distillation Visualisation}
\label{app-sec:distillation-viz}

\begin{figure*}[ht]
\centering
\footnotesize
\includegraphics[scale=0.65]{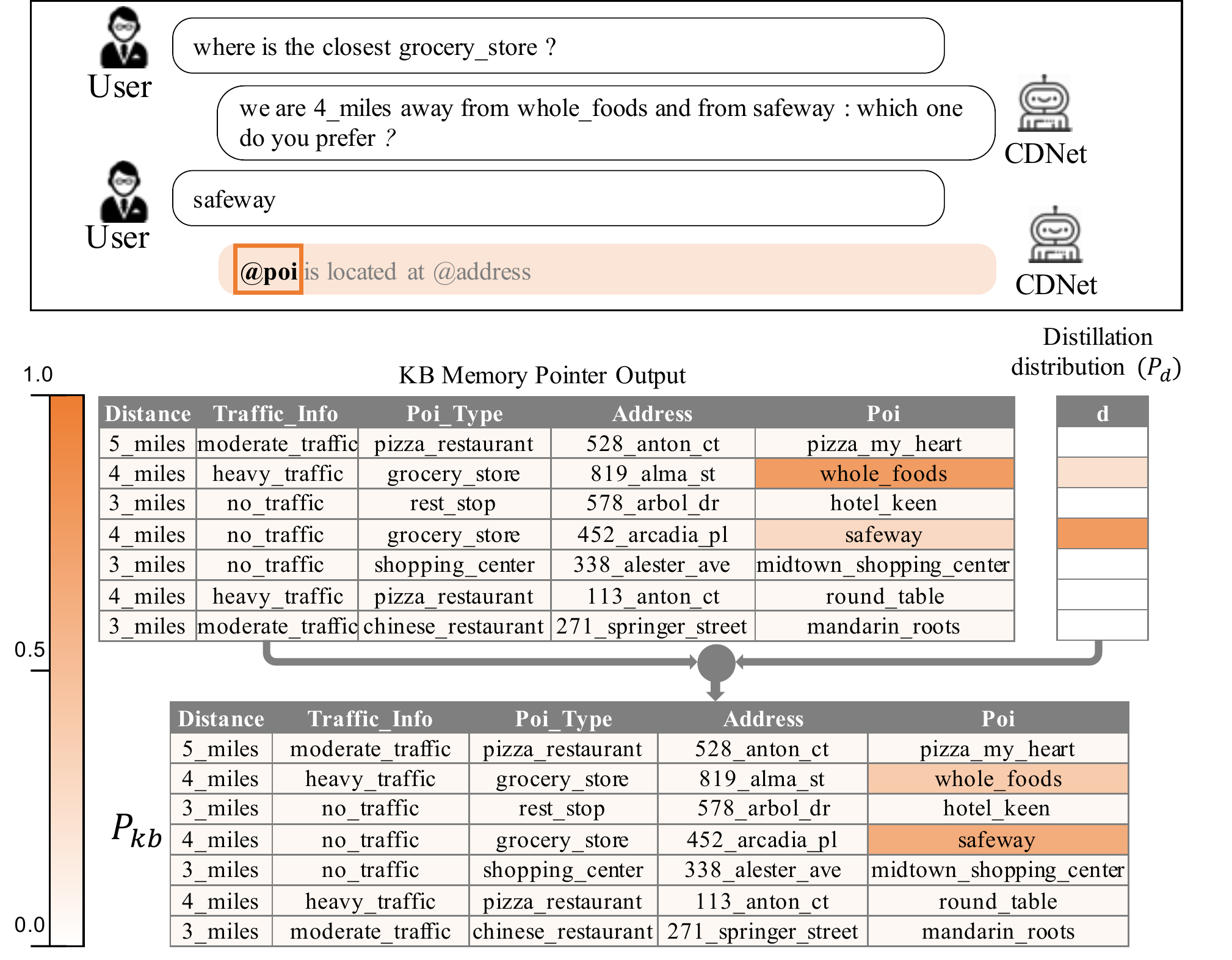}
\caption{Attention visualization of a decode time step of an example from SMD dataset. $P_{kb}$ corresponds to the sketch tag \textit{@poi}. $P_{kb}$ is computed by combining the output of the KB memory pointer and the distillation distribution $P_{d}$.}
\label{fig:appendix_example-one}
\end{figure*}

\begin{figure*}[ht]
\centering
\footnotesize
\includegraphics[scale=0.65]{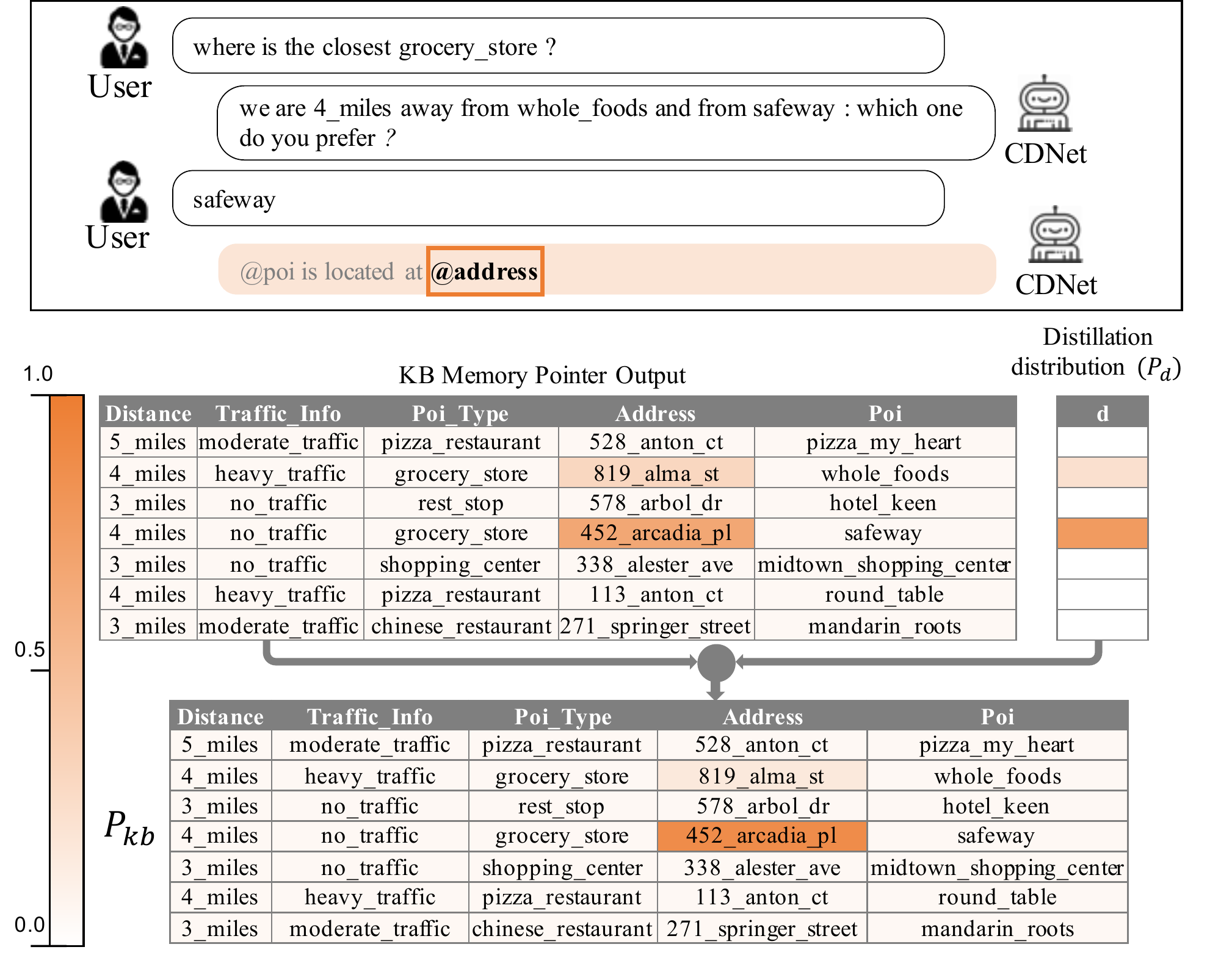}
\caption{Attention visualization of a decode time step of an example from SMD dataset. $P_{kb}$ corresponds to the sketch tag \textit{@address}. $P_{kb}$ is computed by combining the output of the KB memory pointer and the distillation distribution $P_{d}$.}
\label{fig:appendix_example-two}
\end{figure*}

We show the visualisation of how the KB distillation distribution helps the decoder rectify the incorrect KB memory pointer inference in Figure \ref{fig:appendix_example-one}. Figure \ref{fig:appendix_example-two} shows how the KB distillation distribution helps increase the confidence associated with the correct entity in the KB.

\section{Datasets}
We present statistics of SMD, CamRest and WOZ in Table \ref{tab:data}. 

\begin{table}[ht]
\centering
\begin{tabular}{c|c|c|c}
\toprule
& \multicolumn{1}{c|}{\textbf{SMD}} & \multicolumn{1}{c|}{\textbf{CamRest}} & \multicolumn{1}{c}{\textbf{WOZ}} \\
\cmidrule{1-4}
\textbf{Train Dialogs} & 2425 & 406 & 1839 \\
\textbf{Val Dialogs}   & 302  & 135 & 117 \\
\textbf{Test Dialogs}  & 304  & 135 & 141 \\
\bottomrule
\end{tabular}
\caption{Statistics of the three datasets.}
\label{tab:data}
\end{table}

\section{Domain-Wise Results}

Table \ref{tab:app-smd} and Table \ref{tab:app-woz} show the domain wise entity F1 scores of SMD and WOZ datasets respectively. We note that \sys\ either has the best or the second-best performance in domain wise scores.

\begin{table}[t]
\centering
\footnotesize
\begin{tabular}{l|c|c|c|c|c|c}
\toprule
\textbf{Model} & \textbf{BLEU} & \textbf{F1} & \textbf{MSE F1} & \textbf{Cal} & \textbf{Wea} & \textbf{Nav} \\
\midrule
MLM & 17.0 & 54.6 & - & 66.7 & 56 & 46.9 \\
DSR & 12.7 & 51.9 & - & 52.1 & 50.4 & 52.0 \\
Ent. Const. & 13.9 & 53.7 & - & 55.6 & 52.2 & 54.5 \\
TTOS & 17.4 & 55.4 & - & 63.5 & \textbf{64.1} & 45.9 \\
DFNet & 14.4 & 62.7 & - & 73.1 & 57.6 & \textbf{57.9} \\
GLMP & 13.9 & 59.6 & 59.6 & 70.2 &  58.0 & 54.3 \\
EER & 17.2 & 59.0 & 55.1 & 71.8 & 57.8 & 52.5 \\
FG2Seq & 16.8 & 61.1 & 59.1 & 73.3 & 57.4 & 56.1 \\
\midrule
\sys\ & \textbf{17.8} & \textbf{62.9} & \textbf{\textbf{62.9}} & \textbf{75.4} & 61.3 & 56.7 \\
\bottomrule
\end{tabular}
\caption{Domain wise Entity F1 performance of \sys\ and baselines on the SMD dataset.}
\label{tab:app-smd}
\end{table}

\begin{table}[t]
\centering
\footnotesize
\begin{tabular}{l|c|c|c|c|c|c}
\toprule
\textbf{Model} & \textbf{BLEU} & \textbf{F1} & \textbf{MSE F1} & \textbf{Hot}& \textbf{Att}& \textbf{Res}  \\
\midrule
DSR & 9.1 & 30.0 & - & 27.1 & 28.0 & 33.4 \\
DFNet & 9.4 & 35.1 & 34.8 & 30.6 & 28.1 & 40.9 \\
GLMP  & 6.9 & 32.4 & - & 28.1 & 24.4 & 38.4 \\
EER & 13.6 & 35.6 & 35.0 & 35.7 & \textbf{43.0} & 34.2 \\
FG2Seq & 14.6 & 36.5 & 36.0 & 34.4 & 37.2 & 38.9 \\
\midrule
\sys\ & 11.9 & \textbf{38.7} & \textbf{38.6} & \textbf{36.3} & 38.9 & \textbf{41.7} \\
\bottomrule
\end{tabular}
\caption{Domain wise Entity F1 performance of \sys\ and baselines on WOZ dataset.}
\label{tab:app-woz}
\end{table}

\section{Qualitative Example}
\label{app-sec:qual-example}
Table \ref{tab:qualitative-example} shows responses predicted by \sys, EER and FG2Seq for an example from the WOZ dataset.

\section{Human Evaluation} 
\label{app-sec:human-eval}
\begin{figure*}[ht]
\centering
\includegraphics[scale=0.5]{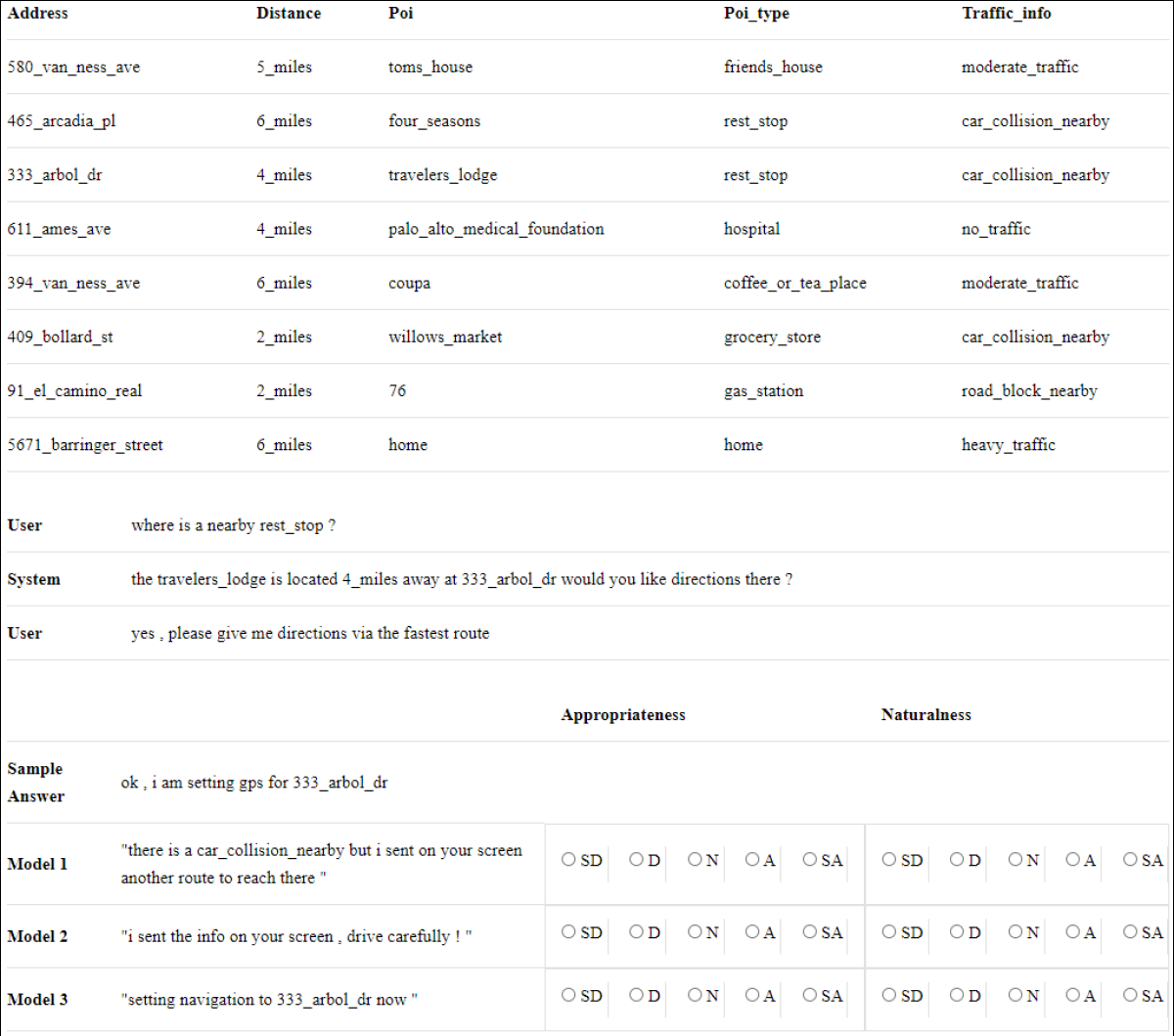}
\caption{A sample human evaluation task used for collecting appropriateness and naturalness of responses generated by three (anonymized) models.}
\label{fig:annotator}
\end{figure*}

Figure \ref{fig:annotator} shows a screenshot of the task used for collecting human judgements.


\begin{table*}
    \centering
    \footnotesize
    \begin{tabular}{c|c|c|c|c|c|c|c}
    \toprule
    \textbf{Address} & \textbf{Area} & \textbf{Choice} & \textbf{Name} & \textbf{Postcode} & \textbf{Pricerange} & \textbf{Ref} & \textbf{Food} \\
    \midrule
   regent\_street\_city\_centre & centre & a\_lot & pizza\_hut\_city\_centre & cb21ab & cheap & oog9kq5v & italian \\    
   4\_-\_6\_rose\_crescent & centre & a\_lot & la\_raza & cb23ll & cheap & oog9kq5v & spanish \\    
   finders\_corner\_newmarket\_road & east & a\_lot & the\_missing\_sock & cb259aq & cheap & oog9kq5v & international \\
   106\_regent\_street\_city\_centre & centre & a\_lot & curry\_garden & cb21dp & expensive & oog9kq5v & indian \\    
    \midrule
    \end{tabular}
    \centering
    \footnotesize
    \begin{tabular}{c|p{13cm}}
    \toprule
    \textbf{usr-1} & i need a \textbf{cheap} \textbf{spanish} restaurant . are there any around ? \\
    \textbf{sys-1} & yes , \textbf{la\_raza} is a restaurant serving \textbf{spanish} food in the \textbf{cheap} price range . it is located at \textbf{4\_-\_6\_rose\_crescent} . would you like a reservation ? \\
    \textbf{usr-2} & i would like a restaurant in the \textbf{centre} . \\
    \textbf{sys-2} & \textbf{la\_raza} is a cheap restaurant serving spanish food in the centre part of town . it ' s address is \textbf{4\_-\_6\_rose\_crescent} and it ' s post code is \textbf{cb2311} . \\
    \textbf{usr-3} & great . can you book me a table for 7 on tuesday at 11:45 and provide me with a reference number ? \\
    \midrule
    \textbf{Gold} & all right , i ' ve made a reservation . your reference number is \textbf{oog9kq5v} . what else can i do for you ?
     \\
    \midrule
    \midrule
    \textbf{EER} & i ' m sorry , but i was unable to book that for you . would you like to try another restaurant ?  \\
    \midrule
    \textbf{FG2Seq} & i ' m sorry , but the \textbf{la\_raza} restaurant restaurants . would you like me to find another restaurant ? \\
    \midrule
    \textbf{\sys\ } & i have booked you at \textbf{la\_raza} . your reference number is \textbf{oog9kq5v} . is there anything else i can help you with ? \\
    \bottomrule
    \end{tabular}
\caption{Example dialog from WOZ along with responses generated by EER, FG2Seq and \sys.}
\label{tab:qualitative-example}
\end{table*}

\end{document}